\documentclass[10pt,twocolumn,letterpaper]{article}

\usepackage[pagenumbers]{cvpr} 

\usepackage{graphicx}
\usepackage{multirow}
\usepackage{makecell}
\usepackage{amsmath}
\usepackage{amssymb}
\usepackage{booktabs}
\usepackage{pifont}
\usepackage[accsupp]{axessibility}  

\usepackage[pagebackref,breaklinks,colorlinks]{hyperref}

\usepackage[capitalize]{cleveref}
\crefname{section}{Sec.}{Secs.}
\Crefname{section}{Section}{Sections}
\Crefname{table}{Table}{Tables}
\crefname{table}{Tab.}{Tabs.}

\def\cvprPaperID{4027} 
\def\confName{CVPR}
\def\confYear{2023}

\begin{document}

\title{Improving Table Structure Recognition with Visual-Alignment Sequential Coordinate Modeling}

\author{Yongshuai Huang\thanks{Equal contribution.} $^{,1}$ \hspace{0.5 cm} Ning Lu$^{*,1}$ \hspace{0.5 cm} Dapeng Chen$^{1}$ \\
Yibo Li$^{2}$ \hspace{0.5 cm} Zecheng Xie$^{1}$ \hspace{0.5 cm} Shenggao Zhu$^{1}$ \hspace{0.5 cm} Liangcai Gao$^{2}$ \hspace{0.5 cm} Wei Peng$^{1}$\\
$^{1}$ Huawei Technologies Ltd. \hspace{0.5cm}$^{2}$ Peking University\\
{\tt\small \{huangyongshuai1,luning12,chendapeng8,xiezecheng1,zhushenggao,peng.wei1\}@huawei.com} \\
{\tt\small \{yiboli,gaoliangcai\}@pku.edu.cn}
}
\maketitle

\begin{abstract}
Table structure recognition aims to extract the logical and physical structure of unstructured table images into a machine-readable format. The latest end-to-end image-to-text approaches simultaneously predict the two structures by two decoders, where the prediction of the physical structure (the bounding boxes of the cells) is based on the representation of the logical structure. However, the previous methods struggle with imprecise bounding boxes as the logical representation lacks local visual information. To address this issue, we propose an end-to-end sequential modeling framework for table structure recognition called \textbf{VAST}. It contains a novel coordinate sequence decoder triggered by the representation of the non-empty cell from the logical structure decoder. In the coordinate sequence decoder, we model the bounding box coordinates as a language sequence, where the left, top, right and bottom coordinates are decoded sequentially to leverage the inter-coordinate dependency. Furthermore,  we propose an auxiliary visual-alignment loss to enforce the logical representation of the non-empty cells to contain more local visual details, which helps produce better cell bounding boxes. Extensive experiments demonstrate that our proposed method can achieve state-of-the-art results in both logical and physical structure recognition. The ablation study also validates that the proposed coordinate sequence decoder and the visual-alignment loss are the keys to the success of our method. 
\end{abstract}

\section{Introduction}
\label{sec:intro}
Tables are an essential medium for expressing structural or semi-structural information. Table structure recognition, including recognizing a table's logical and physical structure, is crucial for understanding and further editing a visual table. The logical structure represents the row-column relation of cells and the spanning information of a cell. The physical structure contains not only the logical structure but also the bounding box or content of the cells, focusing on the exact locations in the image. 

Table recognition can be implemented by an end-to-end encoder-decoder paradigm. Such methods excel at predicting the logical structure but usually produce less accurate physical structures, \emph{i.e.}, bounding boxes of cells or cell contents. However, the bounding box accuracy is essential to downstream tasks, such as text information extraction or table QA. This work designs the sequential coordinate decoding and enforces more visual information to produce more accurate bounding boxes.

\begin{figure}
    \centering
	\begin{subfigure}{0.48\columnwidth}
		\includegraphics[width=0.95\columnwidth]{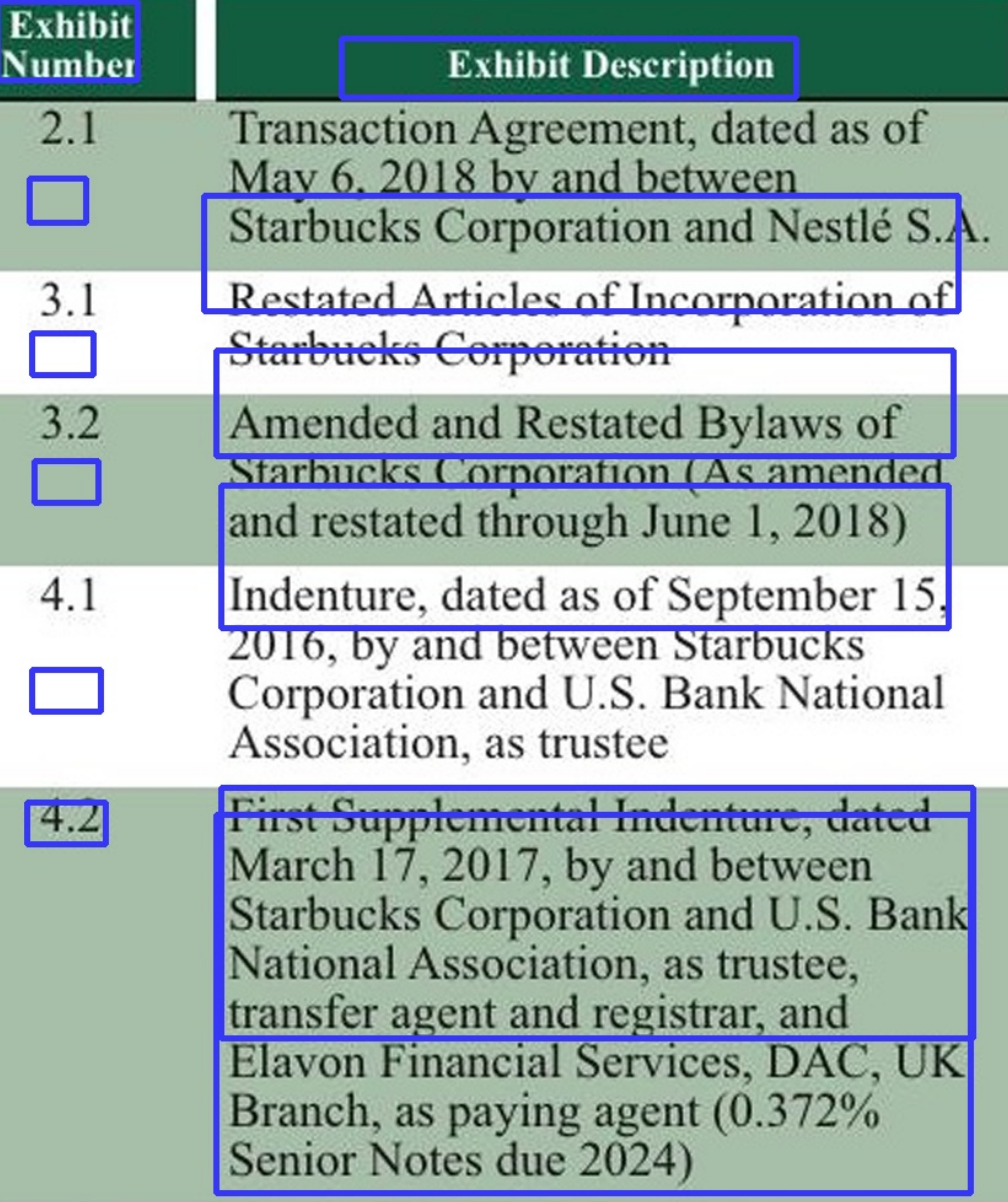}
		\caption{TableFormer (Baseline)}
	\end{subfigure}
	\hfill
	\begin{subfigure}{0.48\columnwidth}
		\includegraphics[width=0.95\columnwidth]{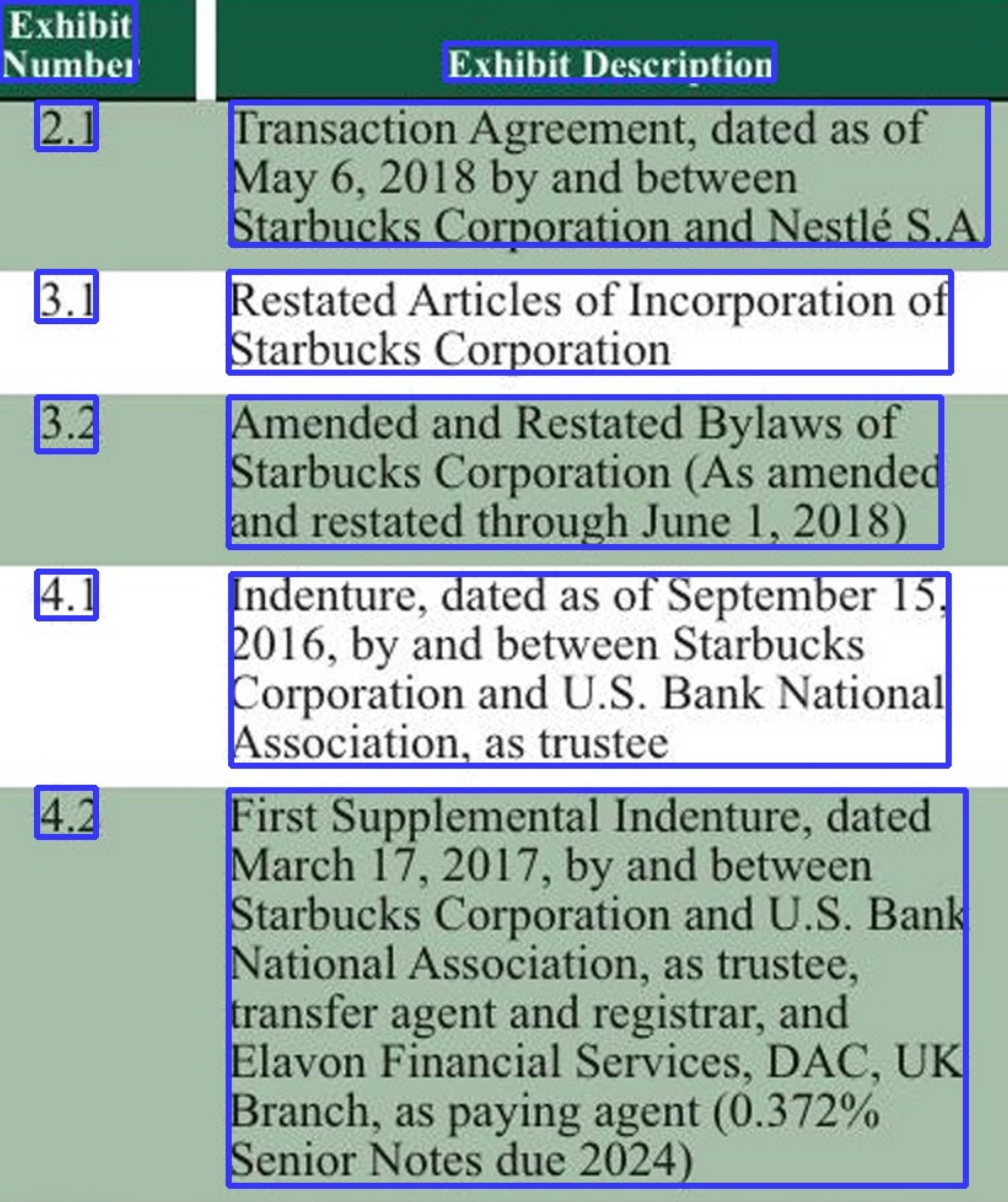}
		\caption{VAST (Ours)}
	\end{subfigure}
	\caption{Visualization comparison of the bounding box predicted by TableFormer and VAST.  Our results are more accurate, which is vital for downstream content extraction or table understanding tasks. The image is cropped from the table with id 7285, which comes from FinTabNet.}
	\label{fig:bbox_compare}
\end{figure}

In the coordinate sequence decoder, the start embedding of the non-empty cell is the representation from the HTML sequence decoder. The representation usually contains a more global context of the table and has fewer local visual details. Because the local visual appearance is vital for predicting accurate coordinates, we align the representation of non-empty cells from the HTML sequence decoder with the visual features from the CNN image encoder. In particular, a visual-alignment loss is designed to maximize the cosine similarity of the 
paired visual-HTML representation in the image. In summary, our contributions are threefold.

\begin{itemize}
    \item We propose a coordinate sequence decoder to significantly improve the table's physical structure accuracy upon an end-to-end table recognition system.
    \item We introduce a visual-alignment loss between the HTML decoder and coordinate sequence decoder. It enforces the representation from the HTML decoding module contains more detailed visual information, which can produce better bounding boxes for the non-empty cells.
    \item We develop an end-to-end sequential modeling framework for table structure recognition, the comparison experiments prove that our method can achieve state-of-the-art performance and the ablation experiments show the effectiveness of our method.

\end{itemize}

\section{Related Work}

The recent deep learning approaches have shown excellent performance on table structure recognition tasks. These methods can be divided into three categories: methods based on splitting and merging, methods based on detection and classification, and image-to-text generation methods.

\noindent\textbf{Methods based on splitting and merging.} These methods consist of two stages. The first stage detects rows and columns, then splits the table into multiple basic text blocks through the intersection of rows and columns; the second stage merges text blocks to restore the structure.

Several works focus on splitting the rows and columns better. For example, DeepDeSRT \cite{Schreiber2017} and TableNet \cite{Paliwal2019} adjusted FCN from the semantic segmentation to segment rows and columns.  DeepTabStR \cite{Siddiqui2019} applied deformable convolution to Faster R-CNN\cite{Ren2015}, FPN\cite{Lin2017}, and R-FCN\cite{Dai2016}, which has a wider receptive field to capture the table line this can split accurate table rows and columns. Khan \etal \cite{Khan2019} and  Li \etal \cite{Li2021} used a bi-directional gated recurrent unit network to identify the pixel-level row and column separators. Inspired by DETR, TSRFormer\cite{TSRFormer} formulated table separation line prediction as a line regression problem and they proposed a separator 
regression transformer to predict separation lines from table images directly. 

Several merging methods have been developed to recognize tables containing cells that span rows or columns. The SPLURGE method\cite{Tensmeyer2019} proposed the idea of table splitting and merging. They designed a merging model to merge cells span multiple columns or rows. To achieve a more accurate merged result, \cite{sem2022} fuse both visual and semantic features to produce grid-level features. RobusTabNet \cite{RobusTabNet}  proposed a spatial CNN-based separation line prediction module to split the table into a grid of cells, and a Grid CNN-based cell merging module was applied to recover the spanning cells. TRUST\cite{TRUST} introduced an end-to-end transformer-based query-based splitting module and vertex-based merging module. The splitting module is used to extract the features of row/column separators, and the row/column features are further fed into the vertex-based merging module to predict the linking relations between adjacent basic cells.

\noindent\textbf{Methods based on detection and classification.} The basic idea of this method is first to detect the cells and then classify the row and column relationships between the cells. A graph can be constructed based on the cell and connection to obtain the table structure. 

For the irregular layout table, a good cell detection result could effectively improve the accuracy of table recognition, \cite{Prasad2020, GTE, Qiao2021, Long2021} were committed to improving the accuracy of cell detection. Some other researchers aimed to classify the cell relationship to construct table structure \cite{Clinchant2018},\cite{Qasim2019}, \cite{Li2020}, \cite{Xue2019}. They utilized ground truth or OCR results to get text blocks. Then they regarded text blocks as vertexes to construct a graph and used the graph-based network to classify the relationship between cells.

The most recent approaches put cell detection and cell relation classification into one network. TableStructNet\cite{Raja2020} and FLAG-NET \cite{Liu2021} both utilized Mask R-CNN\cite{He_2017_ICCV} network to obtain the region of cells and cell visual features. They both utilized the DGCNN architecture in \cite{Qasim2019a} to model the interaction between geometrically neighboring detected cells. Hetero-TSR\cite{Hetero_TSR} proposed a novel Neural Collaborative Graph Machines (NCGM) that leverages modality interaction to boost the multimodal representation for complex scenarios. Lee \etal \cite{Lee2022_planargraphs} formulated tables as planar graphs, and they first obtained cell vertex confidence maps and line fields. After that, they reconstruct the table structure by solving a constrained optimization problem.

\noindent\textbf{Methods based on image-to-text generation.} These methods treat the structure of the table (HTML or latex, etc.) as a sequence, and adopt the end-to-end image-to-text paradigm to recognize the table structure.

Deng \etal \cite{Deng2019} used the classic IM2MAKEUP framework\cite{Deng2017} to recognize the logical structure of the table, where a CNN was designed to extract visual features, and an LSTM with an attention mechanism was used to generate the latex code of the table. Zhong \etal \cite{EDD} tried to generate the logical structure and the cell content with an encoder-dual-decoder (EDD) architecture. In the decoding stage, they used two attention-based recurrent neural networks, one was responsible for decoding the table structure code, and the other was responsible for decoding the content. TableMaster\cite{TableMaster} and TableFormer \cite{tableformer} leveraged the transformer decoder to 
improve the decoder of EDD. In addition, they used the regression decoder to predict the bounding box instead of the content. Since the lack of local visual information, the bounding boxes predicted by these methods were less accurate. In this paper, we treat the bounding box prediction as a coordinate sequence generation task, and cooperate with visual alignment loss to produce more accurate bounding boxes.

\section{Task Definition}
\begin{figure}
	\centering
	\includegraphics[width=0.98\columnwidth]{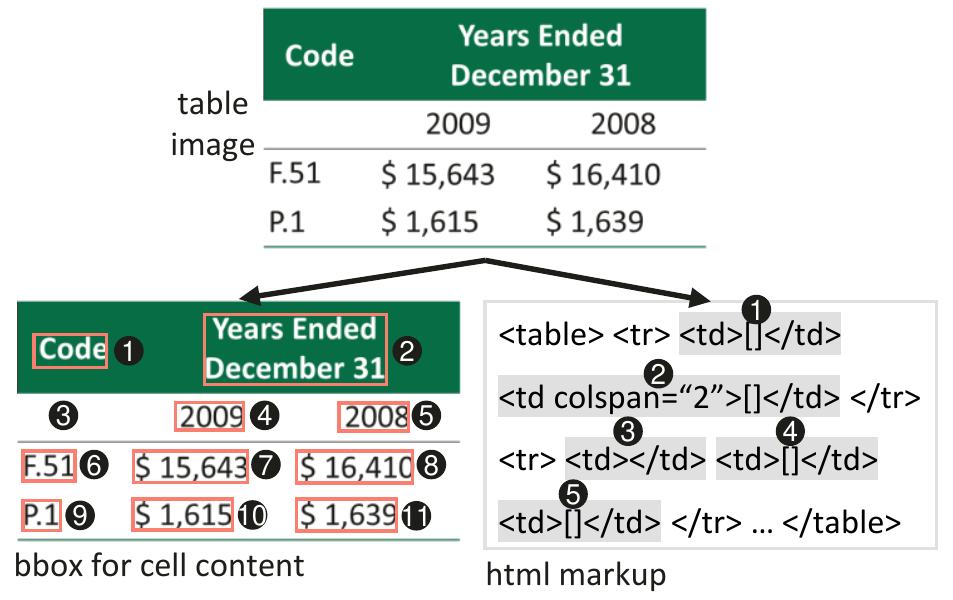}
	\caption{Visualization of table HTML markup and cells. Cell \ding{182} is a spanning cell that spans two columns, and cell \ding{184} is an empty cell with no content. `[]' refers to the content of the cell. }
	\label{fig:markup}
\end{figure}

For a given table image, our goal is to predict its logical structure and physical structure end-to-end. Specifically, the logical structure refers to the HTML of the table, and the physical structure refers to the bounding box coordinates of all non-empty cells. We use $S = [s^{1}, \ldots, s^{T}]$ to indicate the tokenized HTML sequence, where $T$ is the length of sequences and $s$ is a token of predefined HTML tags. We define $B = \{ \mathbf{b}^{1}, \ldots, \mathbf{b}^{N}\}$ is the set of sequences of all non-empty cells, where $\mathbf{b} = ( x_{\text{left}}, \, y_{\text{top}}, \, x_{\text{right}}, \, y_{\text{bottom}} )$, is a sequence of non-empty cell bounding box coordinates and each coordinate is discretized into an integer. An example of HTML for a table and content bounding boxes of non-empty cells is shown in Fig. \ref{fig:markup}.

\section{Methodology}

\begin{figure*}
	\centering
	\includegraphics[width=0.98\textwidth]{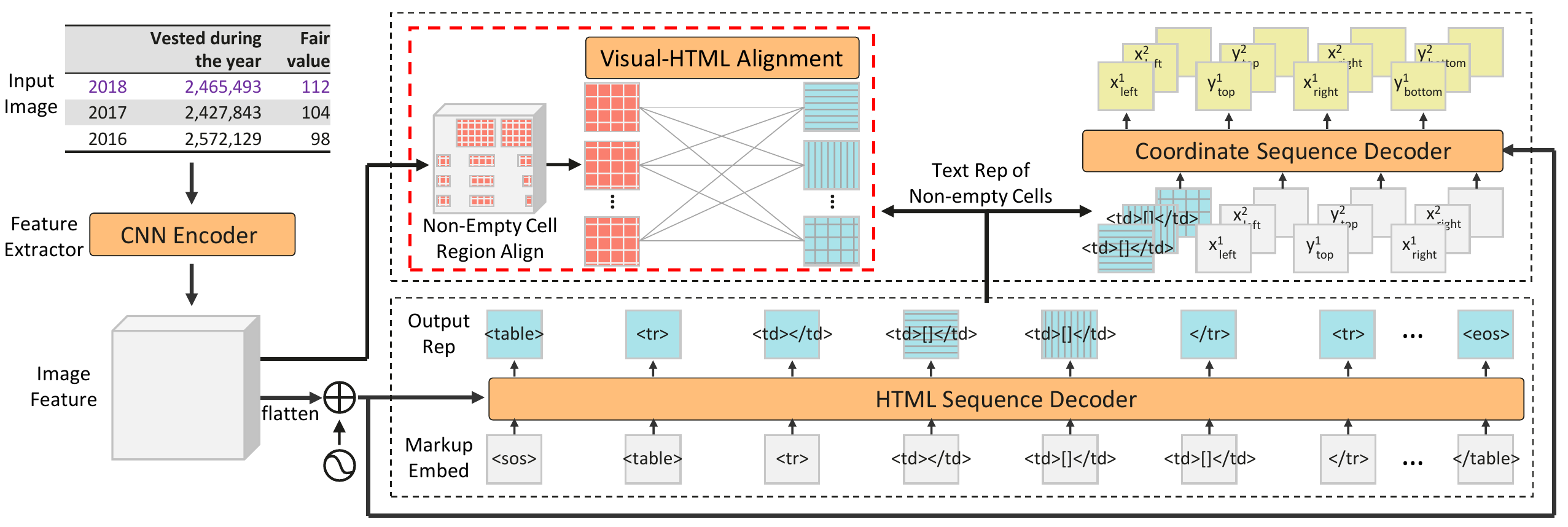}
	\caption{Architecture of our proposed VAST. The red dotted zone refers to the operations only in training.}
	\label{fig:architecture}
\end{figure*}

Our framework consists of three modules:
a CNN image encoder, an HTML sequence decoder and a coordinate sequence decoder. Given a table image, we extract the feature map through the CNN image encoder. The feature map will be fed into the HTML sequence decoder and the coordinate sequence decoder to produce a HTML sequence and bounding boxes of the non-empty cells, respectively. The representation of non-empty cells from the HTML sequence decoder will trigger the coordinate sequence decoder.  
To enforce the local visual information of the representation, visual-alignment loss is employed during training.
The model architecture is illustrated in Fig. \ref{fig:architecture}.

\subsection{CNN Image Encoder}
We use a modified ResNet~\cite{Lu2021} equipped with multi-aspect global content attention as the CNN image encoder. The resulted image feature map is C4, which is from the output of the last convolutional layer of the 4-th stage.
The input of the encoder is a RGB image with a size of $H \times W \times 3$. The output of the encoder is feature map $\mathbf{M}$ with a size 
$\frac{H}{16}\times\frac{W}{16}\times{d}$.

\subsection{HTML Sequence Decoder}

The logical structure of a table contains information such as the number of cells, rows, columns, adjacencies, spanning, etc. In this paper, we use HTML to represent the logical structure of a table. The ground truth HTML of table logical structure is tokenized into structural tokens. As in the work~\cite{TableMaster}, we use merged label to represent a non-spanning cell to reduce the length of HTML sequence. Specifically, we use 
\emph{\textless td\textgreater\textless/td\textgreater} and \emph{\textless td\textgreater[]\textless/td\textgreater} to denote empty cells and non-empty cells, respectively. For spanning cells, the HTML is tokenized to \emph{\textless td}, \emph{colspan=``n"} or \emph{rowspan=``n"}, \emph{\textgreater}  and \emph{\textless/td\textgreater}. We use the first token \emph{\textless td} to represent a spanning cell.

As shown in Fig. \ref{fig:architecture}, the HTML sequence decoder is a transformer with a stack of $N=3$ identical layers. The memory keys and values are the flattened feature map $\mathbf{M}$ added with the positioning encoding. The queries are shifted structure tokens. The output of the transformer is a HTML sequence, which is decoded by auto-regression. The output of the $t$-th step is a distribution: $ p(s_t | \mathbf{M}, s_{1:t-1})$. In training, we employ the cross-entropy loss:
\begin{equation}
    	\mathcal{L}_{s} \!=\! -\log p(S^{*}| \mathbf{M}) \!=\! - \sum_{t=2}^{n} \log p(s_{t}^{*} |s_{1:t-1}^{*}, \mathbf{M}),  \label{loss:s}
\end{equation}
where $S^{*}$ is the ground truth HTML of the target table. The start token $s^{*}_{1}$ or $s_{1}$ is a fixed token \emph{\textless sos\textgreater} in both training and testing phrase.

\subsection{Coordinate Sequence Decoder}

For coordinate prediction, we cascade coordinate sequence decoder after HTML sequence decoder. The decoder is triggered by a non-empty cell $s^{nc}_{i}$. The left, top, right and bottom coordinates are decoded one element at a time. In particular, each of the continuous corner coordinates is uniformly discretized into an integer between $[0, n_{\rm bins}]$. In the decoder, we utilize the embedding of the 
previously predicted coordinates to predict the latter coordinate, which inject contextual information into the prediction of the next coordinate. The procedure of the coordinate sequence decoder is also illustrated in Fig. \ref{fig:architecture}.

Similar to the HTML sequence decoder, the coordinate sequence decoder takes the flattened feature map $\mathbf{M}$ with positioning encoding as memory keys and values, and takes the shifted coordinate tokens as queries. The embedding of the start token is the representation of $s^{nc}_{i}$  from the HTML sequence decoder, denoted by $f^{nc}_{i}$ . The output of the $t$-th step is a distribution: $p(c_{t}| c_{1:t-1}, f^{nc}_{i}, \mathbf{M})$, where $\{c_{t}\}_{t=1}^{4}$ is discrete random variable ranging within $[0, n_{\rm bins}]$. 

The coordinate sequence decoder is also trained by the cross-entropy loss function:

\begin{equation}
    \mathcal{L}_{c}  = -\frac{1}{K} \sum_{i=1}^{K} \sum_{t=1}^{4}\log p(c_{t}^{i*}|c_{1:t-1}^{i*}, f^{nc}_{i}, \mathbf{M}), \label{loss:c}
\end{equation}
where $K$ is the number of non-empty cells in the image, and $c^{i*}_{1}, c^{i*}_{2}, c^{i*}_{3}, c^{i*}_{4}$ correspond to ground truth of the left, top, right and bottom coordinates of $i$-th cell. 
The representation $f^{nc}_{i}$ is from the HTML sequence decoder, it contains contextual information and visual information that can help the coordinate sequence decoder to characterize different cells.

\noindent \textbf{Discussion.} The proposed coordinate sequence decoder is related to Pix2Seq~\cite{pix2seq}. Pix2Seq also decodes the object's coordinates step-by-step but has three distinct differences from our method.
\begin{itemize}
\item Pix2Seq lacks the global information to guide where to decode the coordinates. In contrast, our method predicts the bounding box based on the representation of cells, which are obtained from the table's global contexts. 

\item Pix2Seq needs the sequence augmentation methods to facilitate the training. One of the reasons is that the model needs to perform the classification and the localization simultaneously, which is not easy to converge. While our method doesn't need such an operation.

\item  Pix2Seq decodes different bounding boxes one by one from a long sequence, while our method can perform the bounding box decoding in parallel. This is because we can collect the representation of all non-empty cells, then feed them to the coordinate sequence decoder at once.
\end{itemize}

\subsection{Visual-alignment loss.} To enrich the local visual information in the start embedding of the coordinate sequence decoder, we propose a visual alignment loss to assist the learning of the coordinate sequence decoder. The main motivation of the visual alignment loss is to align the logical structural representation of a non-empty cell with its visual feature. 

During training, we gather the HTML representation of all non-empty cells
$\{f_{i}^{nc}\}_{i=1}^{K}$.  We use ROIAlign\cite{He_2017_ICCV} to extract the visual representation $\{g_{i}^{nc}\}_{i=1}^{K}$ for each non-empty cell. The visual feature is cropped from the image feature map $\mathbf{M}$  according to the bounding box of the ground truth. It is further projected to be a vector having the same dimension with $f_{i}^{nc}$. Given a table image with $K$ non-empty cells, we have $K$ visual-HTML pairs ($g^{nc}_{i}$, $f^{nc}_{i}$). An InfoNCE~\cite{infonce} loss is employed between  $f^{nc}_{i}$ and all visual representation $\{g_{i}^{nc}\}_{i=1}^{K}$ in the image. The visual alignment loss is:
\begin{equation}
    \centering
	\mathcal{L}_{va} = - \sum_{i=1}^{K}{\log(\frac{\exp(f^{nc}_{i} \cdot g^{nc}_{i}/ \tau)}{ \sum_{j=1}^{K}\exp(f^{nc}_{i} \cdot g^{nc}_{j}/ \tau)})}, \label{loss:va}
\end{equation}
where $\tau$ is a temperature hyper-parameter, which is set to 0.04. The cosine similarity is measured by dot product here. It is worth mentioning that this loss is only employed during training, and does not impose any burden on the model during inference.

\subsection{Implementation Details}
We denote our method, \textbf{V}isual-\textbf{A}lignment \textbf{S}equential Coordinate \textbf{T}able Recognizer, as VAST. Some hyper-parameters in the methodology are as follows: The images are resized to $608 \times 608$. The dimension $d$ of the image feature is set to be 512. Both decoders are composed of a stack of $N$ = 3 identical layers, and the number of multi-head $h$ is set to 8. 
The value of $n_{\rm bins}$ is set to 608.

\noindent \textbf{Training.}  The unified loss $\mathcal{L}$ is a combination of losses $\mathcal{L}_{s}, \mathcal{L}_{c}$ and $\mathcal{L}_{va}$ in Eq . \ref{loss:s}, Eq. \ref{loss:c} and Eq. \ref{loss:va}: 

\begin{equation}
    \mathcal{L} = \lambda_1 \mathcal{L}_{s} + \lambda_2 \mathcal{L}_c + \lambda_3 \mathcal{L}_{va}, 
\end{equation}
where $\lambda_1$, $\lambda_1$, and $\lambda_3$ are set to be 1.0, 1.0 and 1.0, respectively. We trained our VAST from scratch using AdamW\cite{AdamW} as the optimizer. The initial learning rate is 1e-4, which decreases by 0.1 per step.  To prevent overfitting, we set the dropout\cite{droupout} rate of the HTML sequence decoder and the coordinate sequence decoder to 0.1. The maximum length for the HTML sequence decoder is set to 500. We trained 48 epochs on 4 Tesla V100 GPUs, and the mini-batch size is 3. The output size of the ROIAlign is  2 $\times$ 2, and we use a linear transformation to project the flattened visual representation $g_{i}^{nc} \in \mathbb{R}^{512 \times 4}$ to a vector 
with the size of 512.

\begin{table*}[t]
	\centering
	\caption{The public datasets for table structure recognition. ``PDF” refers to multiple input modalities, such as images, text, etc., which can be extracted from PDF. ``CAR" indicates cell adjacency relationship. ``Det" indicates the evaluation of detection. ``Cell BBox" and ``Content BBox" refer to the bounding box of cells and content, respectively.  ``IC19B2H' and ``IC19B2M" stand for ``ICDAR2019 TrackB2 historical" and ``ICDAR2019 TrackB2 Modern" respectively.}
	\begin{tabular*}{\hsize}{@{\extracolsep{\fill}}lrrrccccc}
		\toprule
		\multirow{2}{*}{Dataset}    
		& \multicolumn{3}{c}{\#Samples} 
		& \multirow{2}{*}{\makecell[c]{Input \\ Modality}} 
		&  \multirow{2}{*}{\makecell[c]{Cell \\ Content}} 
		& \multirow{2}{*}{\makecell[c]{Cell \\ BBox}}      
		& \multirow{2}{*}{\makecell[c]{Content \\ BBox}} 
		& \multirow{2}{*}{Metric}\\
		\cline{2-4}          
		&  \makecell[c]{Train}  & \makecell[c]{Val} & \makecell[c]{Test} & & & & & \\
		\midrule
		\multicolumn{9}{l}{Logical Structure Recognition} \\  
		TABLE2LATEX-450K\cite{TABLE2LATEX_450K}            &  447K+     & 9,322     & 9,314     & Image   & \ding{51} & \ding{55} & \ding{55} & BLEU \\
		TableBank\cite{tablebank}                    & 130K+     & 10,000     & 5000     & Image   & \ding{55} & \ding{55} & \ding{55} & BLEU \\
		PubTabNet\cite{EDD}                    & 500K+     & 9,115     & 10,000    & Image   & \ding{51} & \ding{55} & \ding{51} & TEDS     \\
		FinTabNet\cite{GTE}                    & 92K     & 10,635   & 10,656   & PDF   & \ding{51} & \ding{55} & \ding{51} & TEDS     \\
		\multicolumn{9}{l}{Physical Structure Recognition} \\
		UNLV\cite{UNLV}                         & -         & -      & 558     & Image   & \ding{55} & \ding{51} & \ding{55} & Det   \\
		ICDAR2013\cite{Goebel2013}                    & -         & -      & 156     & PDF   & \ding{51} & \ding{55} & \ding{51} & CAR  \\
		IC19B2H\cite{Gao2019}                 & -          & -     & 190    & Image   & \ding{55} & \ding{51} & \ding{55} & CAR     \\
		IC19B2M\cite{Gao2019}                 & -          & -     & 145    & Image   & \ding{55} & \ding{55} & \ding{51} & CAR    \\
		SciTSR\cite{scitsr}                      & 12K       & -   &  3,000  & PDF   & \ding{51} & \ding{55} &  \ding{51} & CAR        \\
	    WTW\cite{Long2021}                          & 10K+     & -      &  3,611   & Image   & \ding{55} & \ding{51} & \ding{55} & CAR   \\
		TUCD\cite{TUCD}                        & -      & -   & 4,500  & Image   & \ding{55} & \ding{51} &  \ding{55} & CAR        \\
		PubTables-1M\cite{PubTables1M}                 & 758K+       & 94,959   &  93,834  & PDF   & \ding{51} & \ding{51} & \ding{51} & GriTS             \\
		\bottomrule
	\end{tabular*}
	\label{tab_all_dataset}
\end{table*}

\noindent\textbf{Inference.} In the inference stage,  we use greedy search for the HTML sequence prediction and coordinate sequence prediction. For cell content, if the input modality is PDF, we use the predicted content bounding box to grab content from PDF. If the input modality is an image, we use PSENET\cite{PSENET} and MASTER\cite{Lu2021} to detect and recognize the text and then merge them according to their bounding box. It is noteworthy that we do not make any corrections to the predicted logical structure and physical structure when inserting the content into the cell. 
Supplementary material provides the details of how to fetch the content.

\section{Experiments}

\subsection{Datasets and Evaluation Metrics}
\noindent\textbf{Datasets.}  We investigate the publicly accessible table structure recognition benchmark datasets, as shown in \cref{tab_all_dataset}. 
we evaluate our method on PubTabNet\cite{EDD}, FinTabNet\cite{GTE}, ICDAR2013\cite{Goebel2013}, IC19B2M\cite{Gao2019}, SciTSR\cite{scitsr} and PubTables-1M \cite{PubTables1M}. 
More details of the datasets refer to supplementary materials.

\noindent\textbf{Evaluation metrics.} PubTabNet and FintabNet use tree-edit-distance-based similarity~(TEDS) \cite{EDD} as the evaluation metric. The metric represents the table HTML as a tree, and the TEDS score is obtained by calculating the tree-edit distance between the ground truth and pred trees. Besides TEDS, we also propose S-TEDS, which only considers the logical structure of the table and ignores the cell content.

For ICDAR2013, IC19B2M, and SciTSR, they apply cell adjacency relations (CAR) \cite{Goebel2013} as an evaluation metric. Specifically, it generates a list of horizontally and vertically adjacency relations between true positive cells and their vertical and horizontal neighbors. Then, precision, recall, and F1 score can be calculated by comparing this list with the ground-truth list. The difference is that SciTSR and ICDAR2013 use cell content to 
match predicted cells and ground truth cells, while IC19B2M uses different thresholds of IoU ($\sigma$) to map a predicted cell to a ground truth cell with the highest IoU and IoU $\geq \sigma$. For these three datasets, we will transform the predicted HTML and bounding boxes to a physical structure format.

GriTS was recently proposed by Smock \etal \cite{GriTS} and was first adopted by PubTables-1M. It first represented the ground truth and predicted tables as matrices, and GriTS is computed by the similarity between the most similar substructures of the two matrices. GriTS addresses the evaluation of cell topology, cell content, and cell location recognition in a unified manner.

\subsection{Comparison with the state-of-the-art methods}

\noindent\textbf{Results of logical structure recognition.} As shown in Tab. \ref{tab:fintabnet}, our VAST outperforms all previous methods on FinTabNet and PubTabNet. The difference between S-TEDS and TEDS is mainly due to errors in content recognition or extraction from PDF. For TableMaster, the higher score of TEDS than S-TEDS is because they correct the logical structure by post-processing when fetching cell contents. Compared with the strong baseline TableFormer, VAST improved the S-TEDS score by 1.83\% on FinTabNet and 0.48\%  on PubTabNet. It is worth mentioning that VAST outperforms the TableFormer by improving TEDS from 93.60\% to 96.31\% on PubTabNet. The improvement of the TEDS score is greater than that of S-TEDS, indicating VAST performs better in extracting content. 

We also investigate the performance of cell detection  ($\rm AP_{50}$, MS COCO AP at IoU=.50)  on PubTabNet.  The results in Tab. \ref{tab:cell_det} show that VAST outperforms TableFormer by improving the $\rm AP_{50}$ from 82.1\% to 94.8\%.

\begin{table}[t]
	\centering
	\caption{Comparision on the FinTabNet and PubTabnet. ``PTN + FTN" means training on PubTabNet and finetuning on FinTabNet. }
	\begin{tabular*}{\hsize}{@{\extracolsep{\fill}}lccc}
		\toprule
		\multicolumn{4}{c}{\textbf{FinTabNet}} \\
		\hline
		Methods       & \makecell[c]{Training \\ Dataset} &  S-TEDS &  TEDS  \\
		\midrule
		Det-Base\cite{GTE}              & PTN              & 41.57  & -         \\
		GTE\cite{GTE}                   & PTN + FTN        & 91.02   & -       \\
		EDD\cite{EDD}                   & PTN              & 90.60   & -       \\
		TableFormer\cite{tableformer}   & FTN              & 96.80   &  -      \\
		\textbf{VAST}                   & FTN              & \textbf{98.63}  & \textbf{98.21} \\
		\bottomrule
	    \multicolumn{4}{c}{\textbf{PubTabNet}} \\
		\hline
		TabStructNet\cite{Raja2020}& SciTSR           &                & 90.10      \\
		FLAG-Net\cite{Liu2021}     & SciTSR           & -              & 95.10      \\
		NCGM\cite{Hetero_TSR}      & SciTSR           & -              & 95.40     \\
		GTE\cite{GTE}              & PTN              & 93.01          & -      \\
		RobustTabNet\cite{RobusTabNet}& PTN            & 97.00          & -      \\
		LGPMA\cite{Qiao2021}       & PTN              & 96.70          & 94.60      \\
		SEM\cite{sem2022}          & PTN              & -              & 93.70      \\
		EDD\cite{EDD}              & PTN              & 89.90          & 88.30     \\
		TableMaster\cite{TableMaster}& PTN            & 96.04          & 96.16      \\
		TableFromer\cite{tableformer}& PTN              & 96.75          & 93.60      \\
		TSRFormer\cite{TSRFormer}  & PTN              & \textbf{97.50}          & -         \\
		TRUST\cite{TRUST}& PTN              & 97.10          & 96.20      \\
		\textbf{VAST}      & PTN    & 97.23  & \textbf{96.31} \\
		\bottomrule
	\end{tabular*}
	\label{tab:fintabnet}
\end{table}

\begin{table}
	\centering
	\caption{Comparison of content bounding box detection~(Det) results on PubTabNet.}
	\begin{tabular}{lcc}
		\toprule
		Methods       & Dataset & $\rm AP_{50}$ (\%) \\
        \midrule
		EDD + BBox\cite{tableformer}        & PTN           & 79.2     \\
		TableFormer\cite{tableformer}     & PTN           & 82.1     \\
		\textbf{VAST}   & PTN  & \textbf{94.8}    \\
		\bottomrule
	\end{tabular}
	\label{tab:cell_det}
\end{table}

\noindent\textbf{Results of physical structure recognition.} The results are shown in Tabs \ref{tab:sciTSR}, \ref{tab:icdar2019} and \ref{tab:pubtables}. VAST exceeds  most previous methods and achieves the new state-of-the-art performance on ICDAR2013, IC19B2M and PubTables-1M. 

On SciTSR, VAST achieves the highest precision score of 99.77\% and the second best F1 score of 99.51\%. The recall score of VAST is lower than that of NCGM. This is mainly because some samples in SciTSR have columns beyond the scope of the image. We regard such data as invalid, so our model ignores these mutilated columns during inference. Some visualizations of such data are presented in supplementary materials.   

On ICDAR2013, several methods, such as DeepDeSRT, NCGM, FLAG-Net, etc., were tested on a randomly selected samples from the test set and did not release their split. Thus they are not directly comparable. For the fairness of the comparison, we only compare with methods that report their results on the ICDAR2013 full test dataset. As shown in Tab. \ref{tab:sciTSR}, our VAST outperforms all previous methods with the best F1-score of 96.52\% when trained with FinTabNet and 95.72\% when trained with SciTSR.

\begin{table}[t]
	\centering
	\caption{Comparison of cell adjacency relation~(CAR) score on the SciTSR and ICDAR2013 datasets.}
	\begin{tabular*}{\hsize}{@{\extracolsep{\fill}}lcccc}
		\toprule
		\multicolumn{5}{c}{\textbf{SciTSR}} \\
		\hline
		Methods       & \makecell[l]{Training \\ Dataset} & P~(\%)& R~(\%) & F1~(\%)  \\
		\midrule
		GraphTSR\cite{scitsr}              & SciTSR             & 95.90  & 94.80  & 95.30             \\
		TabStructNet\cite{Raja2020}          & SciTSR             & 92.70  & 91.30  & 92.00             \\
		LGPMA\cite{Qiao2021}                 & SciTSR             & 98.20   & 99.30   & 98.80     \\
		SEM\cite{sem2022}                  & SciTSR             & 97.70   & 96.52   & 97.11     \\
		RobustTabNet\cite{RobusTabNet}          & SciTSR              & 99.40 & 99.10 &99.30 \\
		FLAG-Net\cite{Liu2021}              & SciTSR              & 99.70 & 99.30 &99.50 \\
		NCGM\cite{Hetero_TSR}                  & SciTSR              & 99.70 & \textbf{99.60} & \textbf{99.60} \\
		TSRFormer\cite{TSRFormer}                  & SciTSR              & 99.70 & \textbf{99.60} & \textbf{99.60} \\
		\textbf{VAST}        & SciTSR             & \textbf{99.77} & 99.26 & 99.51 \\
		\bottomrule
		\multicolumn{5}{c}{\textbf{ICDAR2013}} \\
		\hline
		GraphTSR\cite{scitsr}            & SciTSR              & 88.50  & 86.00  & 87.20 \\
		TabStructNet\cite{Raja2020}       & SciTSR              & 91.50 & 89.70   &90.60 \\
		CycleCenterNet\cite{Long2021}    & WTW                 &  \textbf{95.50} & 88.30   &91.70\\
		LGPMA\cite{Qiao2021}              & SciTSR               & 93.00 & 97.70   &95.30 \\
		GTE\cite{GTE}                & FTN                 & 92.72 & 94.41   &93.50 \\
		\textbf{VAST}    & SciTSR            & 93.84 & \textbf{97.68}   &\textbf{95.72} \\
		\textbf{VAST}    & FTN      & 95.29 & \textbf{97.79}   & \textbf{96.52} \\
		\bottomrule
	\end{tabular*}
	\label{tab:sciTSR}
\end{table}

 \begin{table}[t]
	\centering
	\caption{Comparison of cell adjacency relation~(CAR) F1-score~(\%) on the IC19BM. ``IC19~\dag" refers to the manually annotated ICDAR2019 dataset in CascadeTabNet\cite{Prasad2020}.}
	\begin{tabular*}{\hsize}{@{\extracolsep{\fill}}lcccc}
		\toprule
		\multirow{2}{*}{Methods}       & \multirow{2}{*}{\makecell[c]{Training \\ Dataset}} & \multicolumn{2}{c}{IoU} & \multirow{2}{*}{WAvg.F1}  \\
		\cline{3-4}
		&                                       & 0.5           & 0.6           &               \\
		\midrule
		NLPR-PAL\cite{Gao2019}           & -               & -             & 36.5           & 36.5          \\
		CascadeTabNet\cite{Prasad2020}     & IC19~\dag        & -             & 43.8           & 43.8          \\
		GTE\cite{GTE}               & FTN              & 54.8          & 38.5           & 45.9          \\
		\textbf{VAST}     & FTN              & \textbf{66.8} & \textbf{51.7} & \textbf{58.6} \\
		\bottomrule
	\end{tabular*}
	\label{tab:icdar2019}
\end{table}

On IC19B2M,  we report the results with the IoU thresholds of 0.5 and 0.6 as the competitive baseline method GTE \cite{GTE}. The \textbf{WAvg.F1} score is the weighted average value of F1 scores under each threshold. 
As shown in \cref{tab:icdar2019}, VAST achieves the highest F1-score at the IoU threshold of 0.5 and 0.6, outperforming GTE by 12\% and 13.2\%, respectively. Compared  with  CascadeTabNet, when the IoU threshold is set to 0.6, VAST surpasses it by 7.9\%, even though it used their own labeled ICDAR2019 dataset for training. Inherently, for the overall average F1~(WAvg.F1), VAST achieves the best score of 58.6\%.

PubTables-1M is the most challenging benchmark dataset with 93834 samples for evaluation. As shown in Tab. \ref{tab:pubtables}, we report the results on $\rm Acc_{Cont}$, $\rm GriTS_{Top}$, $\rm GriTS_{Cont}$ and $\rm GriTS_{Loc}$. 
The scores of VAST in $\rm Acc_{Top}$, $\rm GriTS_{Top}$, $\rm GriTS_{Cont}$ are 90.11\%, 99.22\% and 99.14\% respectively, achieving the current state-of-the-art performance. The $\rm GriTS_{Loc}$ score of VAST is lower than that of DETR because DETR uses the bounding box of the content contained in the cell to adjust the predicted bounding box of the cell.

\begin{table}
	\centering
		\setlength\tabcolsep{1pt}
	\caption{Comparison of GriTS~(\%) score on PubTables-1M}
	\begin{tabular*}{\hsize}{@{\extracolsep{\fill}}lcccc}
		\toprule
		Methods      & $\rm Acc_{Cont}$ & $\rm GriTS_{Top}$ & $\rm GriTS_{Cont}$ & $\rm GriTS_{Loc}$ \\
        \midrule
		FasterRCNN\cite{tableformer}                            & 10.39     & 86.16     & 85.38  & 72.11\\
		DETR\cite{tableformer}                & 81.38     & 98.45     & 98.46  & \textbf{97.81}\\
		\textbf{VAST}   & \textbf{90.11}    & \textbf{99.22} & \textbf{99.14}  & 94.99\\
		\bottomrule
	\end{tabular*}
	\label{tab:pubtables}
\end{table}

\begin{figure}
    \centering
	\begin{subfigure}{0.48\columnwidth}
		\includegraphics[width=\columnwidth]{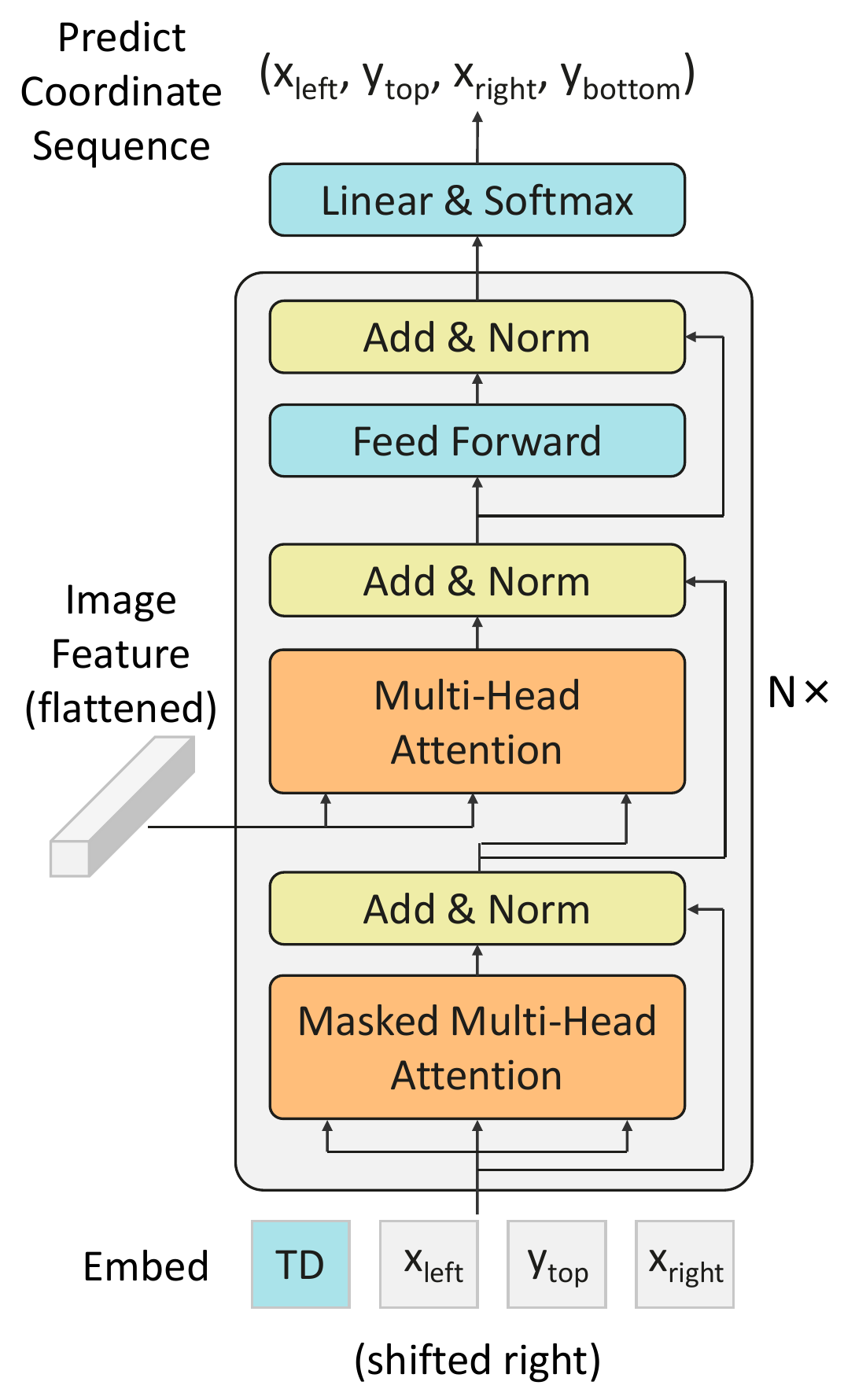}
		\caption{CSD}
	\end{subfigure}
	\hfill
	\begin{subfigure}{0.48\columnwidth}
		\includegraphics[width=\columnwidth]{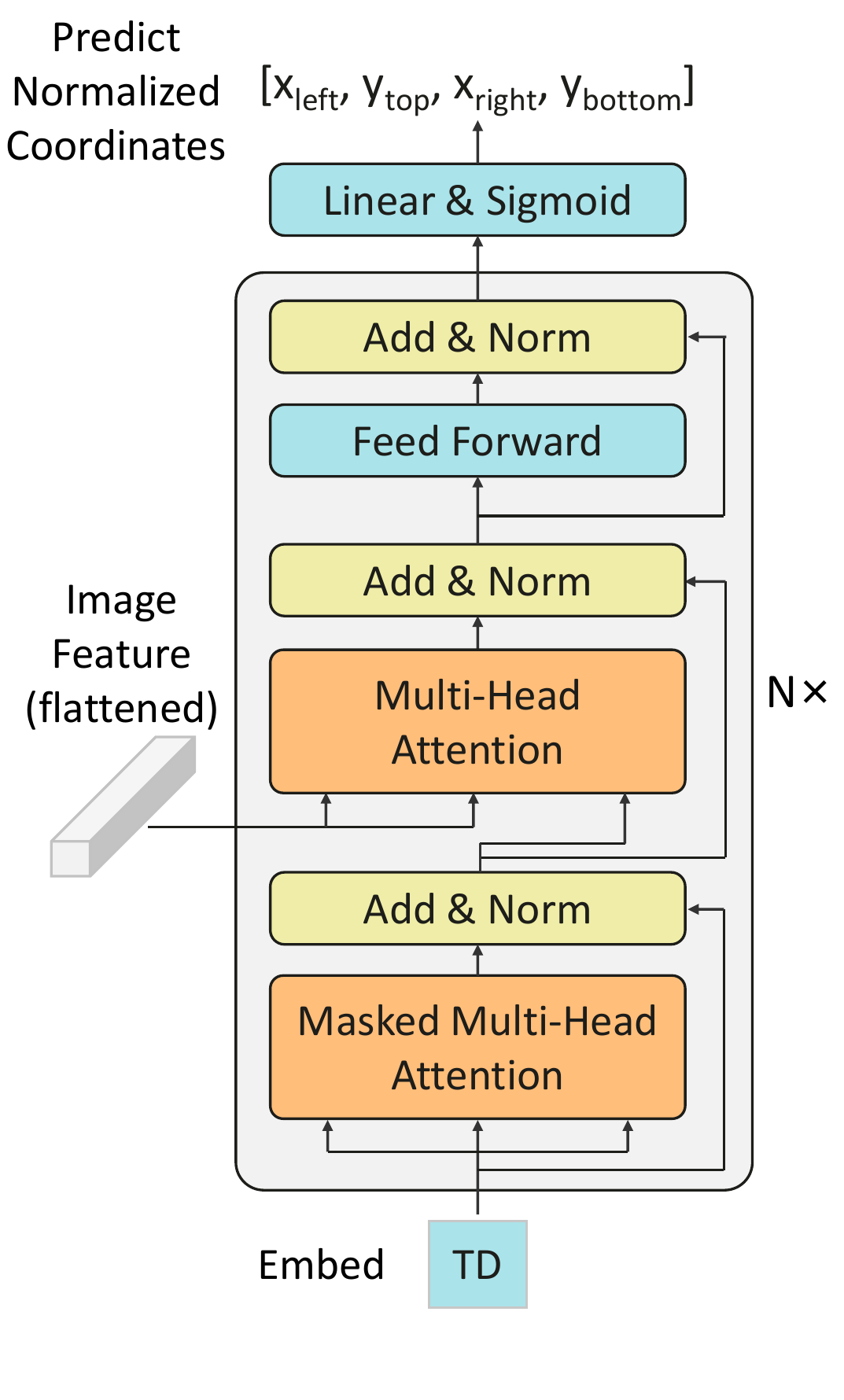}
		\caption{RD}
	\end{subfigure}
	\caption{Architecture of Coordinate Sequence Decoder~(CSD) and Regression Decoder~(RD). `TD' indicates the representation of the non-empty cell from HTML Sequence Decoder. To simplify, position encoding is omitted.}
	\label{fig:ablation_csd_rd}
\end{figure}

\subsection{Ablation Study}

We conduct a set of ablation experiments to verify the effectiveness of our proposed modules. We use FinTabNet for training, and then test on the FinTabNet test set and IC19B2M. The results are in Tab. \ref{tab:ablation01}, where the S-TEDS scores for logical structure and detection AP (MS COCO AP at IoU=.50:.05:.95) and WAvg.F1 scores for non-empty cells are reported.

\begin{table}
	\centering
	\caption{Ablation studies for structure recognition on FinTabNet test set and IC19B2M. ``RD" and ``CSD" indicate regression decoder and coordinate sequence decoder, respectively. ``VA'' refers to visual alignment loss.}
	\begin{tabular*}{\hsize}{@{\extracolsep{\fill}}c|ccc|cc|c}
		\hline
		\multirow{2}{*}{Exp}        & \multicolumn{3}{c|}{Modules} &  \multicolumn{2}{c|}{FinTabNet} & IC19B2M        \\
		\cline{2-7} 
        & RD           & CSD            &  VA             &  S-TEDS   & AP & WAvg.F1  \\
		\hline
		\#1  & \ding{51}        &               &              & 98.22     & 87.3  & 42.5  \\
		\#2  &                  & \ding{51}    &               & 98.48            & 95.6            & 52.1    \\
		\#3  &                 &  \ding{51}     & \ding{51}    & 98.63            & 96.2            & 58.6  \\
		\hline
	\end{tabular*}
	\label{tab:ablation01}
\end{table}

\begin{figure*}

    \centering
	\begin{subfigure}{0.98\textwidth}
		\includegraphics[width=\columnwidth]{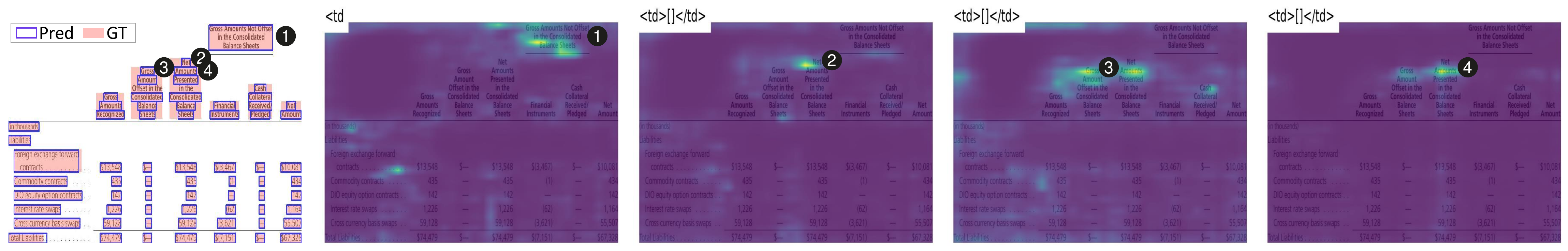}
		\caption{VAST training without VA~(Exp~\#2)}
	\end{subfigure}
	\begin{subfigure}{0.98\textwidth}
		\includegraphics[width=\columnwidth]{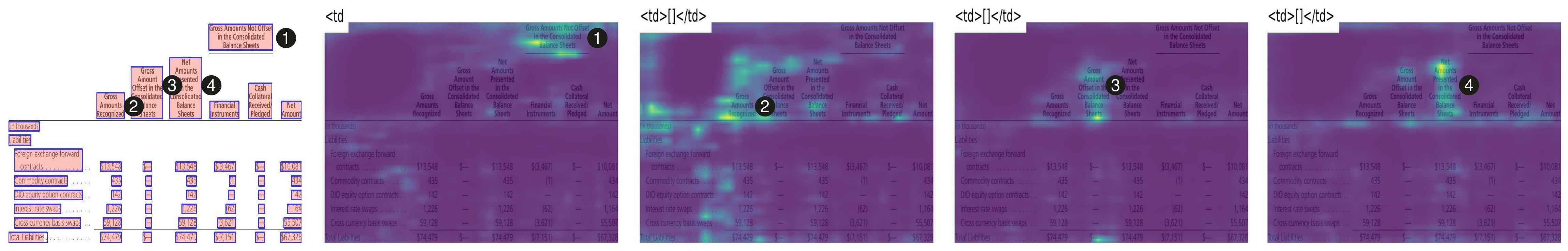}
		\caption{VAST training with VA~(Exp~\#3)}
	\end{subfigure}
	\caption{Visualization of the HTML sequence decoder's cross attention map when predicting the first 4 non-empty cells. This is a hard sample cause the content of the second row of the table is misaligned.}
	\label{fig:ablation}
\end{figure*}

\noindent\textbf{Effectiveness of coordinate sequence decoder.} To validate the effectiveness
of the Coordinate Sequence Decoder~(CSD), we follow TableFormer\cite{tableformer} and TableMaster\cite{TableMaster} to implement a Regression Decoder~(RD) module, 
as shown in \ref{fig:ablation_csd_rd}. 
The difference between the CSD and RD lies in the output header and loss function: 
1) By using a Softmax activation function, CSD generates the discrete coordinate sequence ($x_{\text{left}}, \, y_{\text{top}}, \, x_{\text{right}}, \, y_{\text{bottom}}$) one element at a time, which can consume the previously generated coordinate as additional input when generating the next. RD uses the Sigmoid activation function to output the normalized coordinates of [$x_{\text{left}}, \, y_{\text{top}}, \, x_{\text{right}}, \, y_{\text{bottom}}$] at once. 2) We employ the cross-entropy loss to train the CSD and L1 loss to train the RD.

We use Coordinate Sequence Decoder~(CSD) in Exp~\#2 to learn and predict the content bounding box of the non-empty cell while using the Regression Decoder~(RD) in Exp~\#1, the result are shown in Tab. \ref{tab:ablation01}. On FinTabNet, Exp~\#2 improves the S-TEDS and detection AP score by 0.22\% and 8.3\%. On IC19B2M, Exp~\#2 outperforms Exp~\#1 in WAvg.F1 score by 9.4\%. 
It can be seen that CSD can substantially improve the performance of physical structure recognition, indicating that CSD explicitly models the dependencies among the coordinates and predicts more accurate bounding boxes.

\noindent\textbf{Effectiveness of visual-alignment loss.}
According to Exp~\#3 and Exp~\#2 in Tab. \ref{tab:ablation01}, adding the visual alignment loss~(VA) during training is beneficial to the logical and physical structure recognition. On FinTabNet, Exp~\#3 improves 0.15\% and 0.6\% in terms of the S-TEDS and the AP, respectively. On IC19B2M, Exp~\#3 also gives a gain of 6.5\% for WAvg.F1. 
These results demonstrate the effectiveness and generality of our proposed visual alignment loss.

Furthermore, we show some qualitative results to indicate how VA can enforce the local visual information of the representation of non-empty cells.
In Fig. \ref{fig:ablation}, we visualize the cross-attention~(average over heads and layers) when the HTML sequence decoder predicts the token of non-empty cells. One can see that, when predicting the token of the first non-empty cell, the attention of both models can correctly focus on the cell. When predicting the second non-empty cell, the attention of the model w/o VA  incorrectly focuses on the top text block in the row instead of the entire cell. While the model w/ 
VA can correctly concentrate around the second non-empty cell. The same phenomenon also appeared for the third and fourth non-empty cells. 
Details of cross-attention map generation and more visualizations are presented in the supplement.

\section{Limitations}
Although extensive experimental results demonstrate the effectiveness of our proposed method, there are still two limitations of our proposed VAST. 1) Since our model employs 
an auto-regressive manner and only generates one token at a time during inference, the inference speed is slower than methods based on splitting-and-merging. Specifically, the runtime of VAST, EDD and TRUST are 1.38, 1 and 10 FPS, respectively. 2) We use HTML sequence to represent the logical structure of the table, even if we merge some tags to reduce the length of the HTML sequence, there is still too much redundancy in the sequence, such as `\textless/tr\textgreater', `\textless/td\textgreater', \etc, resulting in higher computation and memory consumption.

\section{Conclusion}

We proposed an end-to-end sequential modeling framework for table structure recognition. This model consists of two cascaded transformer decoders to generate the HTML sequence of the whole table and the coordinates of non-empty cells, respectively. The representation of the non-empty cells from the HTML sequence decoder is used as the start embedding to trigger the coordinate sequence decoder. Besides, we also proposed an auxiliary visual alignment loss, which lets the start embedding of 
each non-empty cell contains more local visual information and then produce a more accurate bounding box. Experimental results have demonstrated that our method has achieved new SOTA performance on several benchmark datasets.

{\small

\bibliographystyle{ieee_fullname}
\bibliography{main}
}

\clearpage
\newpage
\appendix
\onecolumn
\section{Supplement}

\subsection{Datasets}
Among all public accessible datasets, TABLE2LATEX-450K\cite{TABLE2LATEX_450K}, TableBank\cite{tablebank}, PubTabNet\cite{EDD} and FinTabNet\cite{GTE} are used for logical structure recognition, where their logical structures are represented by markup languages such as HTML or Latex. The others are for physical structure recognition, and their structure is described by the bounding box and logical location of the cells, \emph{i.e.} star-row, end-row, start-column, and end-column. Since VAST predicts the logical structure of the table and bounding box of the content, datasets without content bounding box annotations such as TABLE2LATEX-450K, TableBank, UNLV\cite{UNLV}, IC19B2H\cite{Gao2019}, WTW\cite{Long2021} and TUCD\cite{TUCD} are not suitable for our method.

\subsection{Qualitative Results}

\begin{figure*}[ht]
    \centering
    \begin{subfigure}{0.95\textwidth}
		\includegraphics[width=0.98\columnwidth]{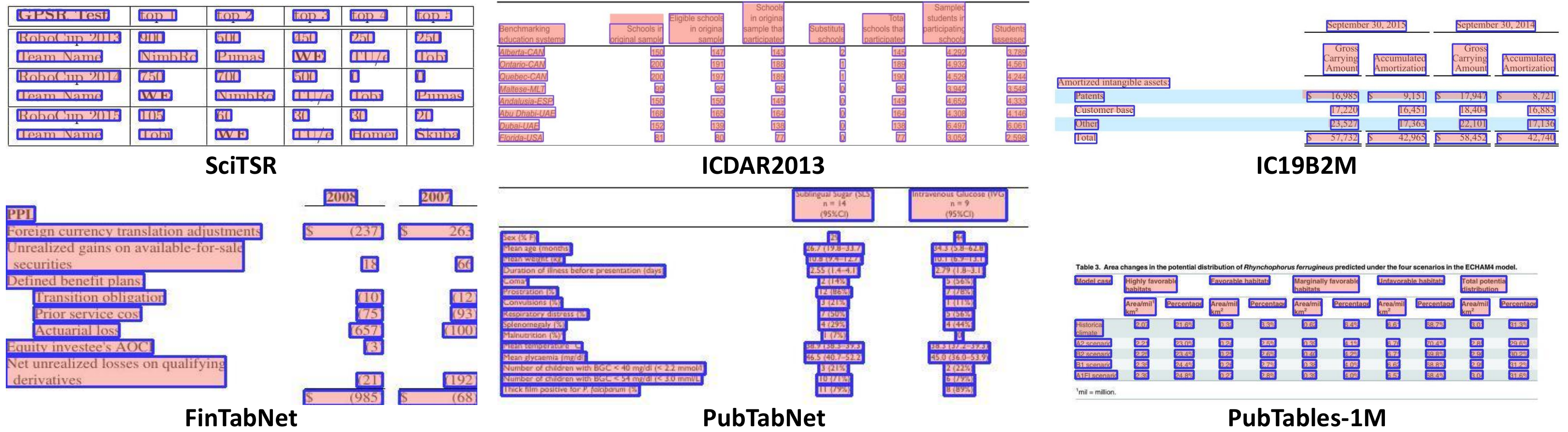}
		\caption{Positive samples}
	\end{subfigure}
	\hfill
	\begin{subfigure}{0.95\textwidth}
		\includegraphics[width=0.98\columnwidth]{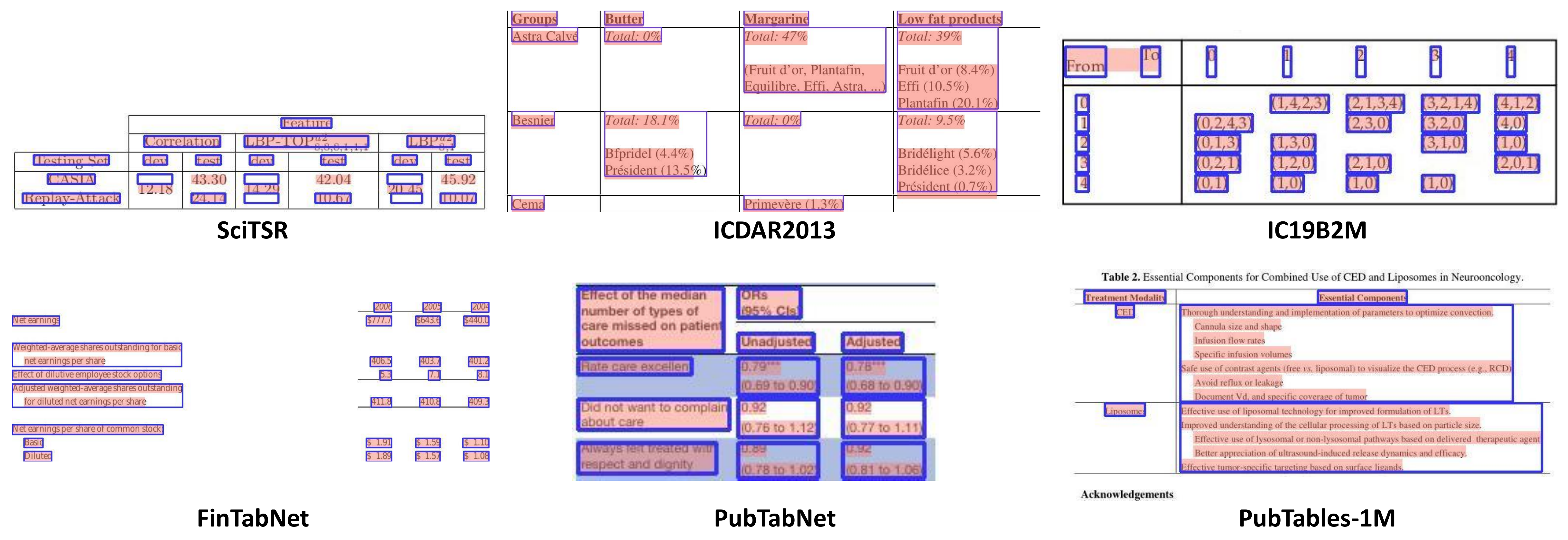}
		\caption{Negative samples}
	\end{subfigure}
	\caption{Visualization of non-empty cell detection results of VAST}
	\label{fig:qualitative_samples}
\end{figure*}

We present some positive and negative samples of the detection results of non-empty cells by VAST in Fig. \ref{fig:qualitative_samples}. From these qualitative results, we can see that the bounding box predicted by VAST can tightly enclose the contents of the cell. Negative samples consist  mainly of tables with over-segmented or over-merged content. There are two reasons for these errors, one is due to the ambiguity of the annotations (samples of FinTabNet and ICDAR2013), and the other is due to the lack of semantic information of cell content (samples of PubTabNet and PubTables-1M).
It is worth noting that, even though VAST incorrectly predicts some cells, it takes local visual information into account when predicting cell bounding boxes. Compared with the results of VAST w/o VA in Fig. 5 of the paper, VAST can significantly reduce over-segmented cells.

\subsection{Details of content extraction}
\begin{figure*}[h]
    \centering
	\includegraphics[width=0.95\textwidth]{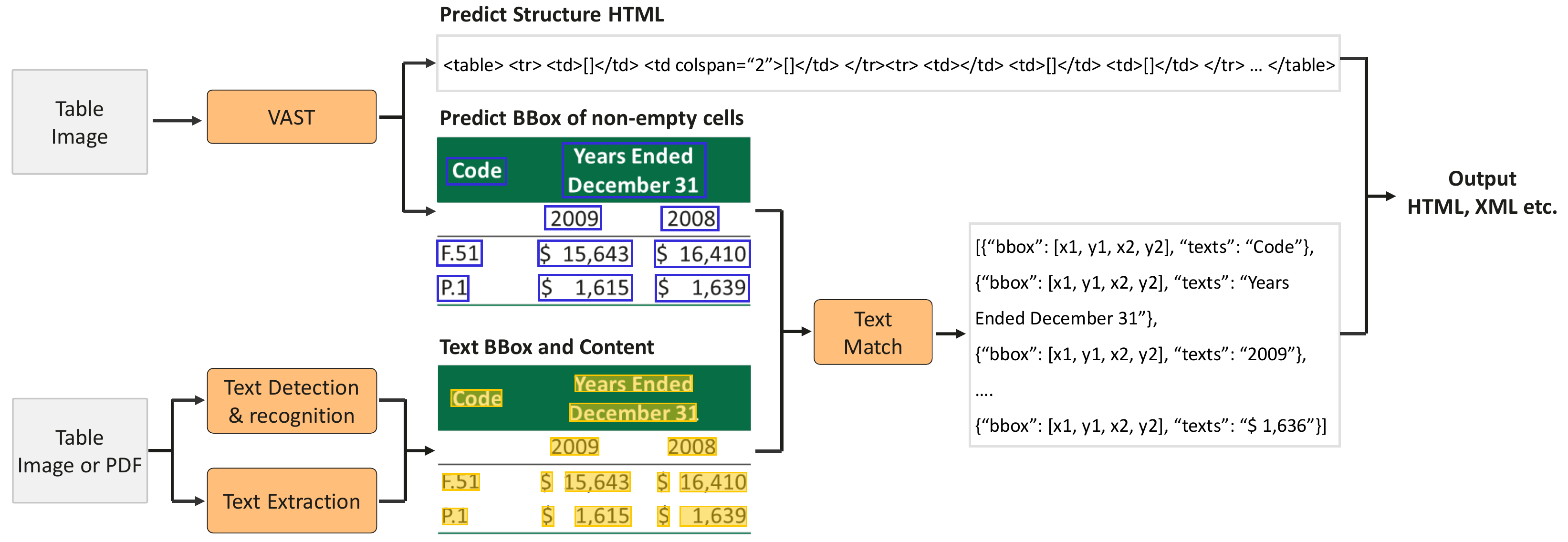}
	\caption{Pipeline of post-processing to obtain content of non-empty cells}
	\label{fig:pipeline}
\end{figure*}

The output of VAST are the logical structure~(HTML) of the table and the bounding box of all non-empty cells, which is incomplete  for evaluation metrics considering the content, so we need to obtain the content of each non-empty cell through simple post-processing. Fig. \ref{fig:pipeline} illstrates the complete pipeline:
\begin{enumerate}
    \item 
    If there is a PDF file of the table, such as FinTabNet, SciTSR, ICDAR2013 and PubTables, we use PDFMiner\footnote{https://www.unixuser.org/~euske/python/pdfminer/index.html}  to extract the content bounding box and content of each text line within the table area from the PDF document. If there are only images of the table, such as PubTabNet, we use PSENET to detect text lines and MASTER to recognize texts in text lines. For the fairness of the comparison, we use the pretrained PSENET and MASTER model of TableMaster\footnote{https://github.com/JiaquanYe/TableMASTER-mmocr}. 
    \item We match text lines with non-empty cells by using the highest IoU and IoU $\geq$ 0.1.
    \item For cells that contain multiple text lines, we sort the text lines left-to-right and top-to-bottom then merge their texts.
\end{enumerate}

After getting the contents of non-empty cells, we combine them with structure HTML to output the HTML or XML result for evaluation.

\subsection{Details of cross-attention weight visualization}
\noindent\textbf{Details of the generation of cross-attention visualization maps.}
In the HTML sequence decoder, we compute the dot-products of the query with all keys, divide each by $\sqrt{d_{k}}$ and apply a Softmax to get the weight of cross-attention. At the $l$-th step of decoding, we collect the cross-attention weight  $ \text{Attns} \in  \mathbb{R}^{h \times 1 \times L}$ 
of each decoder layer to obtain the cross attention weights $ \text{Attns}^{l} = [\text{Attn}^{l}_{1}, \text{Attn}^{l}_{2}, \text{Attn}^{l}_{3}]$ of this step, where $h$ is the number of multi-head and $L$ refers to the size of the flattened image feature. If the token predicted at step $l$ represents a non-empty cell, 
that is, the token is `\textless td\textgreater []\textless/td\textgreater' or  `\textless td', we average the weights over all layers and all heads to get the averaged weight $ \text{Attn}^{l}_{mean} \in \mathbb{R}^{L}$. The averaged weight is reshaped to the size of $\sqrt{L} \times \sqrt{L}$, and the values are normalized to 0-1 and then scaled to 0-255. Finally, the weight map is resized to the size of the image and then overlaid on the original image with 0.8 transparency.

\begin{figure}
    \centering
	\includegraphics[width=0.9\columnwidth]{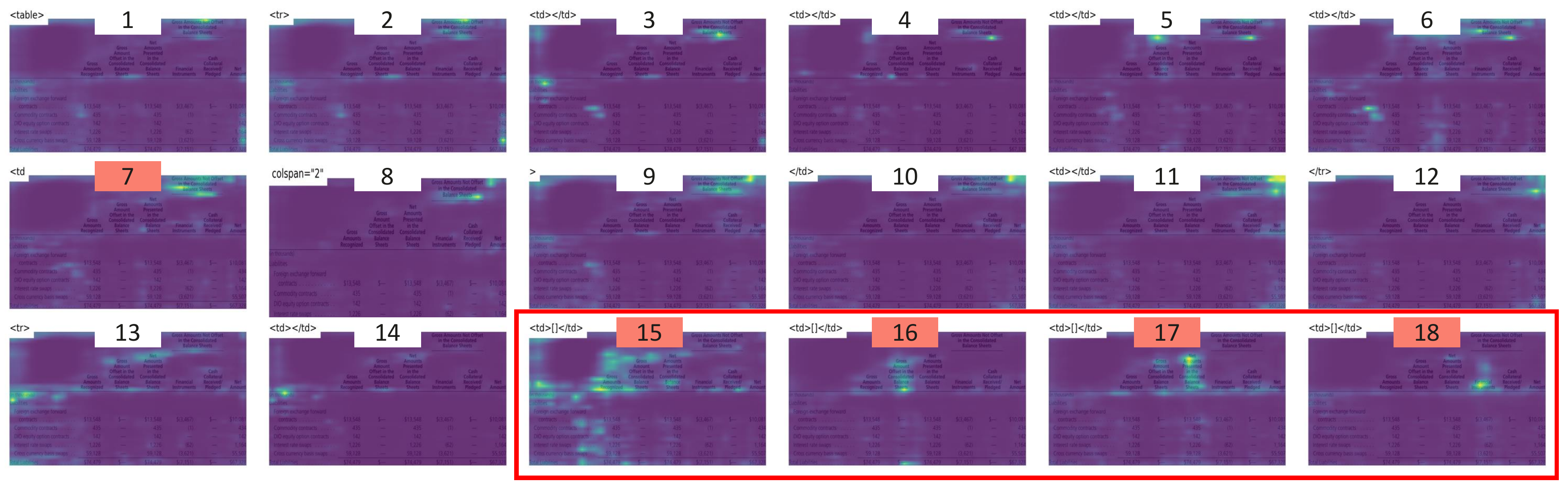}
	\caption{Visualization of cross attention maps of VAST trained \textbf{with} visual-alignment loss.}
	\label{fig:attns_w_va}
\end{figure}

\begin{figure}
    \centering
	\includegraphics[width=0.9\columnwidth]{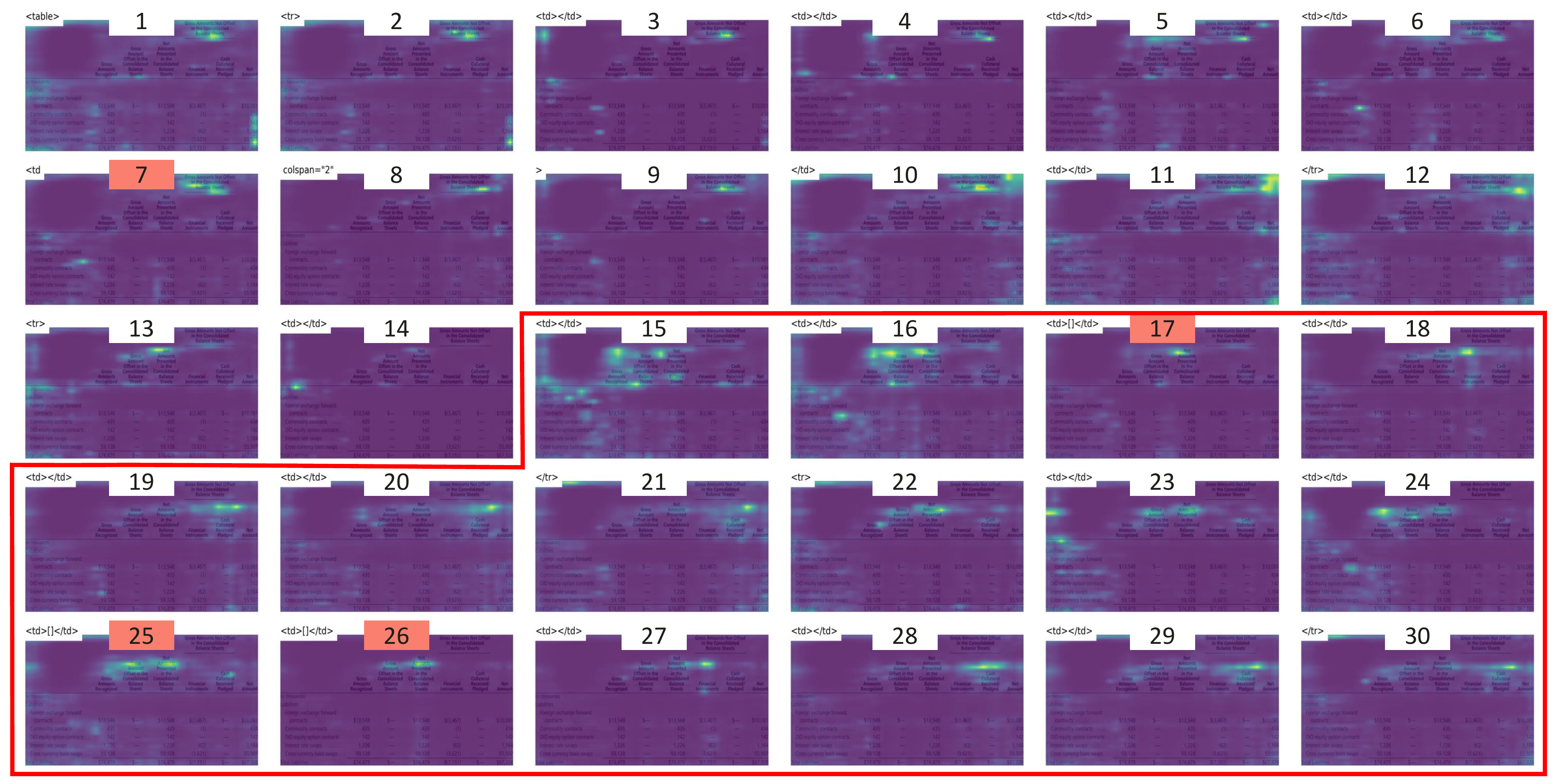}
	\caption{Visualization of cross attention maps of VAST trained \textbf{without} visual-alignment loss.}
	\label{fig:attns_wo_va}
\end{figure}

\noindent\textbf{Comparison of cross-attention visualizations of VAST w/ VA and VAST w/o VA.} Fig. \ref{fig:attns_w_va} and Fig. \ref{fig:attns_wo_va} show cross attention maps of each step as the model predicting the first four non-empty cells. A numeric label with a colored background indicates that the token decoded at this step represents a non-empty cell. 
Obviously, in the first 14 steps, the logical structure results predicted by the two models are consistent, and the attention maps are also almost similar. The difference occurs in step 15, the model trained with VA loss has more attention near the cell text, so it can correctly predict the cell. However, the model trained without VA loss erroneously focuses on the blank space above the text and incorrectly predicts that cell as a blank cell. After step 15, the difference increases. VAST w/ VA can correctly predict the next three cells at steps 16, 17, and 18, however, due to the 
error of step 15 and lack of local visual information,  VAST w/o VA incorrectly predicts a lot of over-segmented cells.

\subsection{Samples with mutilated columns in SciTSR}
\begin{figure}[!ht]
    \centering
	\begin{subfigure}{0.45\columnwidth}
		\includegraphics[width=0.98\columnwidth]{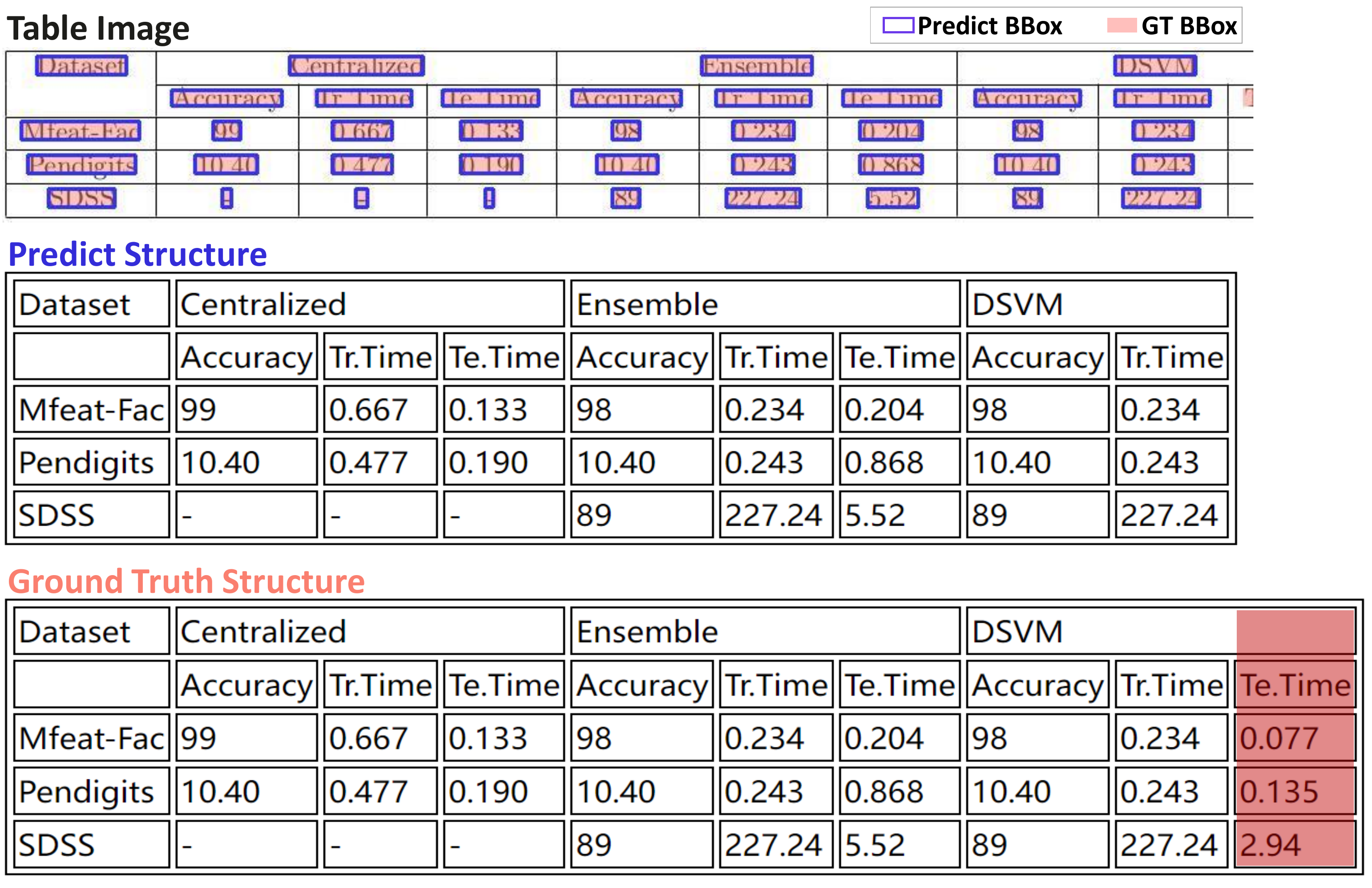}
		\caption{image id:1512.00664v1.7}
	\end{subfigure}
	\begin{subfigure}{0.45\columnwidth}
		\includegraphics[width=0.98\columnwidth]{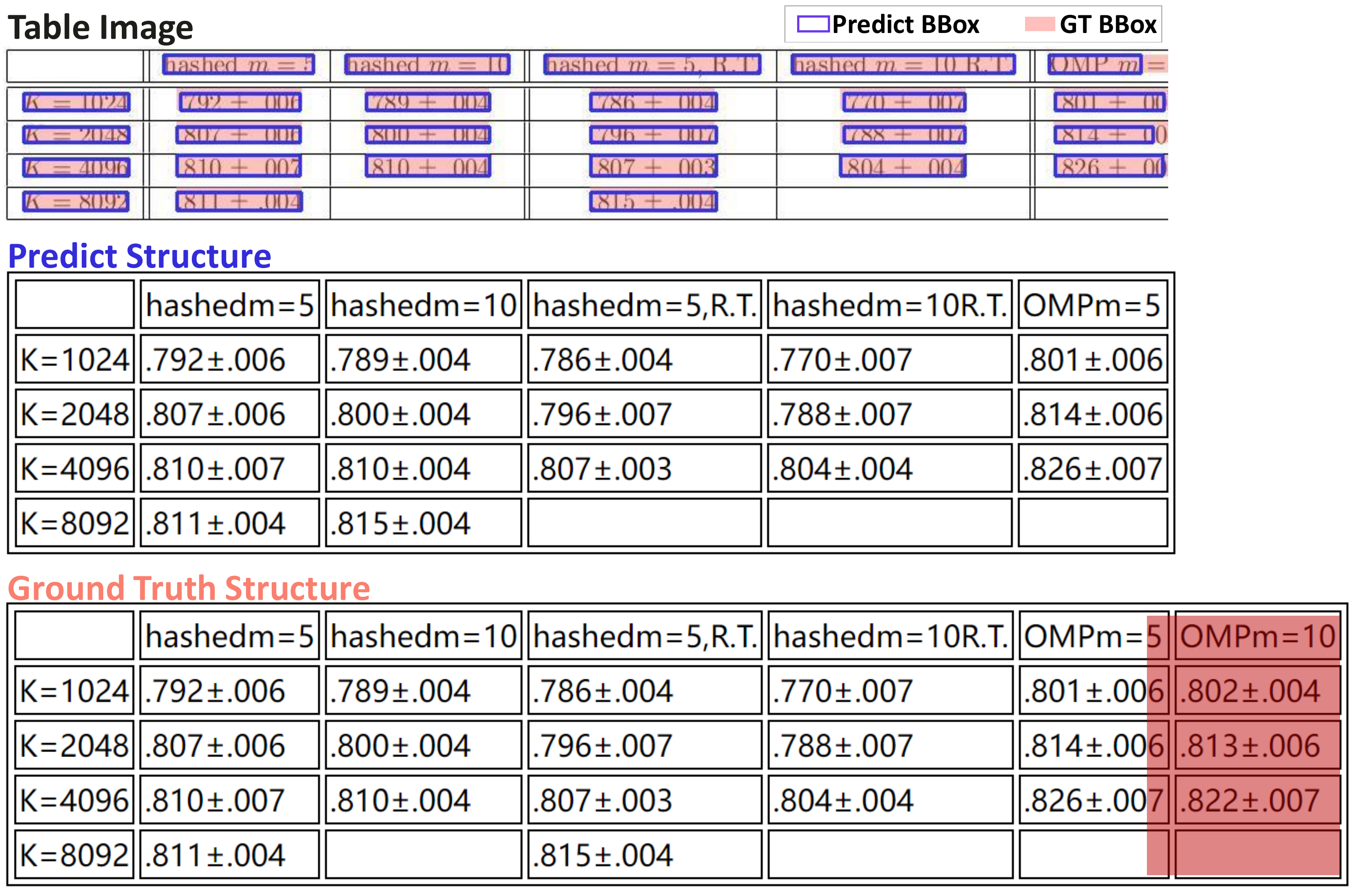}
		\caption{image id:1202.6384v1.2}
	\end{subfigure}
	\caption{Example with mutilated columns in SciTSR. Masked regions in the ground truth are not shown in the image.}
	\label{fig:scitsr}
\end{figure}

Some samples in SciTSR with incomplete columns are shown in Fig. \ref{fig:scitsr}. In the table image, it can be seen that the predicted bounding box matches the ground truth accurately except the mutilated columns. In addition, the structure predicted by VAST is also consistent with the image. However, as you can see from the structure of the ground truth, there are some columns that do not exist in the image, which are highlighted by the mask. This results in a lower recall score for our predictions.

\end{document}